\documentclass[conference]{IEEEtran}
\IEEEoverridecommandlockouts
\usepackage{amsmath,amssymb,amsfonts}
\usepackage{graphicx}
\usepackage{xcolor}
\usepackage{float}
\usepackage{tabularx}
\usepackage{multirow} 
\usepackage{booktabs}
\usepackage{subcaption}
\usepackage{makecell}
\usepackage[colorlinks=true, allcolors=blue]{hyperref}

\def\BibTeX{{\rm B\kern-.05em{\sc i\kern-.025em b}\kern-.08em
    T\kern-.1667em\lower.7ex\hbox{E}\kern-.125emX}}
\begin{document}

\title{Evaluation and Optimization of Rendering Techniques for Autonomous Driving Simulation}

\author{
\IEEEauthorblockN{Chengyi WANG}
\and
\IEEEauthorblockN{Chunji XU}
\and
\IEEEauthorblockN{Peilun WU}
\thanks{Chengyi WANG, Chunji XU, Peilun WU are with the Southern University of Science and Technology, Shenzhen, 518055, China}
}

\maketitle

\begin{abstract}
In order to meet the demand for higher scene rendering quality from some autonomous driving teams (such as those focused on CV), we have decided to use an offline simulation industrial rendering framework instead of real-time rendering in our autonomous driving simulator. Our plan is to generate lower-quality scenes using a game engine, extract them, and then use an IQA algorithm to validate the improvement in scene quality achieved through offline rendering. The improved scenes will then be used for training.
\end{abstract}

\section{Introduction}
Computer graphics rendering technology is essential for autonomous driving simulation. Specifically, when engineers attempt to run a simulation, they must first create a 3D model that specifies how geometries should be placed in the virtual world \cite{lan2016development1,lan2016development2}. 

The image rendering technology used by modern autonomous driving simulators has been relatively complete from the algorithm level. However, due to the limitation of calculation time, the effect of real-time rendering is often unsatisfactory. In order to improve the quality of model rendering used in our autonomous driving laboratory, it is necessary and feasible to use offline rendering instead of real-time rendering to breakthough this time limit. We expect that offline rendering will perform better than real-time rendering. And currently, most rendering tools are real-time rendering. To perform offline rendering, we need to render ourselves, and after offline rendering, we also need to evaluate the rendering effect to see if offline rendering is really better than real-time rendering.

We will first render a scene in real-time, and then render a similar scene offline. Then we will evaluate the rendering effects of the two scenes. We will evaluate the rendering effect in two ways. The first method is to use object detection algorithms to detect rendered targets and see if it can improve the detection effect. Another method is to use the IQA algorithm to evaluate rendered images to see if the quality of offline rendered images is higher than that of real-time rendered images.

The result is in two parts. 
In detection algorithms, offline rendering is slightly better than online ones, and in image quality assessments the advantage is much greater.

\section{Background Research}
\subsection{Building Real-time Simulation Environment}
Real-time refers to the ability of a system to process and respond to data in a timely manner, typically within a few milliseconds or less. In the context of a real-time simulator for autonomous driving, this means that the simulator is able to simulate the driving environment and the behavior of the autonomous vehicle in real-time, responding to inputs and changes in the environment as quickly as possible \cite{xu2019online}.

To highlight the advantage of our offline rendering method in producing high-quality scenes, we will first build a real-time autonomous driving simulation environment, extract appropriate scenes from it, render them offline, and then compare the quality of the renders.

Engineering a simulation environment requires the same features and toolsets used in creating other types of rich interactive content: lighting and physics, particle and weather systems, animation, machine learning \cite{Unity2020,lan2021learning}.

A number of real-time simulators for autonomous driving have been created, including CARLA, Airsim, Udacity Simulator, Gazebo, LGSVL Simulator, Carsim and so on. Among them, CARLA and Airsim are based on a famous game engine Unreal Engine.

\subsection{Precomputation-Based Rendering methods} 
In 2001, Basri and Jacobs showed that reflection from a curved surface, lit by an environment map, could be seen as a spherical convolution of the incident illumination and the reflective properties or BRDF of the surface. \cite{Basri2003}\cite{Ravi2007}

Sloan et al. introduced the term precomputed radiance transfer and led to greatly increased interest in precomputation-based relighting methods. They use full environment maps for lighting instead of discrete light sources. The first innovation was to build on the spherical harmonic methods, but precompute the shadowing and interreflection components, encapsulated in a transfer function on the object surface. \cite{Sloan2002}\cite{Ravi2007}

\subsection{IQA for Rendering}
A method known as CVRKD-IQA proposed by Yin et al. \cite{CVRKD_2022} is especially suitable for our underground parking lot rendering research. 
Knowledge distillation is used for feature extraction from pictures and training the NAR-student agent.
Under the verification of various data sets and algorithms, the results of the IQA algorithm have sufficient credibility.

Bosse et al. \cite{WaDIQaM_2018} implemented an FR-IQA and NR-IQA method by Deep Neural Networks named WaDIQaM.
The pre-study and training of different feature fusion strategies make the DNN system effective on FR and NR IQA scenes.

Hossein and Peyman \cite{NIMA_2018} introduced a neural image assessment algorithm trained on both aesthetic and pixel-level quality datasets. 
This CNN-based NR-IQA method is called NIMA. 
NIMA is a pre-trained IQA algorithm and is popular for effectively predicting the distribution of quality ratings, rather than just the mean scores.

\subsection{Evaluate image quality}

After using the previously mentioned methods for rendering, we will use other methods and metrics to verify the image quality.

\subsubsection{PSNR (Peak Signal-to-Noise Ratio) and MSE (Mean Square Error)}
Peak Signal to Noise Ratio (PSNR) is a commonly used indicator for evaluating image quality. It is determined by calculating the ratio between the maximum possible pixel value in the image and the mean square error caused by noise in the image. Usually, the higher the PSNR, the better the image quality. However, it should be noted that PSNR does not always accurately reflect the human eye's perception of image quality, as the human eye has different sensitivities to different types and intensities of distortion \cite{lan2016action}. Therefore, when evaluating image quality, other factors need to be considered, such as structural similarity index (SSIM) and subjective visual quality assessment \cite{HOULT197671}.

Mean error refers to the difference between each data point in a set of data and the average value of that set of data. It can be calculated by adding up the difference between each data point and the average value and dividing it by the number of data points. 
Mean error is usually used to evaluate the accuracy and accuracy of a set of measurement results, with smaller mean errors indicating that the measurement results are closer to the true values \cite{HOULT197671}.

The MSE formula is as follows:
\begin{equation}
MSE = \sum_{i=1}^{M}\sum_{j=1}^{N}|R(i, j) - F(i, j)|^{2}
\end{equation}

Where R is the reference image, F is the image to be evaluated. And the PSNR formula is as follows:
\begin{equation}
PSNR = 20log_{10}MAX_{p} - 10log_{10}MSE
\end{equation}

Where $MAX_p$ is the maximum possible pixel value in an image (usually 255).
This algorithm has the advantage of convenient computation, but its evaluation results are significantly different from the intuitive perception of the human eye, as the human visual system is more sensitive to brightness information than chromaticity information.

\subsubsection{SSIM(Structural Similarity Index)}
SSIM is an abbreviation for Structural Similarity Index, which is an objective evaluation indicator used to measure image quality. It can compare the structural similarity and differences in brightness, contrast, structure, and other aspects between two images to evaluate their similarity. The higher the SSIM index, the more similar the two images are. The SSIM index is widely used in the field of image processing, such as image compression, denoising, and enhancement.\cite{6059504}

The SSIM formula is as follows:
\begin{equation}
SSIM = [l(x,y)c(x,y)s(x,y)]
\end{equation}
where x and y represent two images, while l(x,y), c(x,y), and s(x,y) represent brightness similarity, contrast similarity, and structural similarity, respectively. Specifically, their calculation formula is as follows:
\begin{equation}
l(x,y) = \frac{2\mu_x\mu_y+C_1}{\mu_x^2 +\mu_y^2+C_1}
\end{equation}

\begin{equation}
c(x,y) = \frac{2\sigma_xy+C_2}{\sigma_x^2 +\sigma_y^2+C_2}
\end{equation}

\begin{equation}
s(x,y) = \frac{\sigma_{xy}+C_3}{\sigma_x\sigma_y+C_3}
\end{equation}

Where $\mu_x$, $\mu_y$ represents the average value of image x and image y, respectively, $\sigma_x$, $\sigma_y$ represents their standard deviations, respectively, $\sigma_{xy}$ represents their covariance. $C_1$, $C_2$, and $C_3$ are constants used to avoid instability when the denominator is 0 or close to 0.

Although the results of the SSIM algorithm are more in line with the human eye's perception, it requires that the two images be of the same size and undergo graying during evaluation. 
In addition, due to the highly nonlinear nature of the human visual system, there are differences between the evaluation results and the actual visual perception.

\subsubsection{NIQE (Natural Image Quality Evaluator)}
The natural image quality evaluator evaluates the quality of an image by extracting quality perception features from each image block and fitting them into a multivariate Gaussian model. Finally, the distance between the original image and the multivariate Gaussian model of the test image is used as the quality of the test image. This framework has been proposed with some methods that mainly focus on natural scene statistics (NSS) feature extraction. In the Natural Image Quality Evaluator (NIQE), Mittal et al. used mean subtraction and contrast normalization (MSCN) fitting parameters as quality perception features. These features are also fundamental features of subsequent work, as they are very effective.\cite{WU202117}

The NIQE formula is as follows:
\begin{equation}
Q(T) = \frac{1}{N}\sum(\frac{||x-\mu||^2}{\sigma^2})
\end{equation}

Where N is the total number of blocks in the test image, $\mu$ and $\sigma^2$ are the mean and variance of the multivariate Gaussian model. This formula calculates the average Euclidean distance between the test image and the multivariate Gaussian model, which is used to evaluate the quality of the test image.

The NIQE algorithm can perform well in super-resolution reconstruction tasks where other algorithms cannot perform well, but it also requires more complex calculations and longer computational time.

\subsection{Object Detection Algorithms} 
Object detection algorithms are computer vision techniques that enable machines to identify and locate objects within an image or video \cite{lan2018real,xiang2016uav}. 
These algorithms use various techniques, including machine learning, deep learning, and computer vision, to detect objects within an image or video and draw bounding boxes around them \cite{lan2019evolving}.

There are generally two main types of object detection algorithms: two-stage detectors and one-stage detectors \cite{lan2022vision}. Two-stage detectors follow a two-step process. The first stage generates a set of potential object proposals, which are regions in the image that may contain objects. These proposals are then refined and classified in the second stage. One popular example of a two-stage detector is the Faster R-CNN (Region-based Convolutional Neural Network) algorithm, which uses a region proposal network (RPN) to generate proposals and then classifies and refines them using subsequent stages. One-stage detectors perform object detection in a single pass without the explicit proposal generation step. These algorithms directly predict the bounding boxes and class labels for objects in an image \cite{lan2022class,gao2021neat}. Single Shot MultiBox Detector (SSD) and You Only Look Once (YOLO) are examples of popular one-stage detectors. SSD divides the image into a grid and predicts multiple bounding boxes and class probabilities at each grid location, while YOLO divides the image into a grid and predicts bounding boxes and class probabilities directly.

Compared to two-stage detectors, one-stage detectors(especially YOLO) are more efficient and provide better performance in detecting small objects. These advantages make them more suitable for application in the field of autonomous driving, and that is the reason why we choose YOLOv3, YOLOv8, and Detectron2 to evaluate our results.

\section{Research methods}
\subsection{Build scene from ground}
For our experimental purposes, we need to use a rendering model based entirely on offline industrial renders. In order to ensure the same work as the recognition algorithm, we select components with the same functions as Carla Town to build the scene. 


For reasons of realism, aesthetics or consistency with the scene, some parameters can be adjusted before using these models.
We combine the models to get the required scene, and then select a suitable perspective in the scene to get a photo for the algorithm to identify in both online and offline engines.

\subsection{Rendering engine and optimize}
To optimize the rendering result, we use an offline Cycles engine to render images of possible autonomous driving scenes.

Cycles is a ray tracing engine in which many post-processing functions are built in. The main feature of ray tracing engines is that the ray tracing engine determines the position of the object first, while the rasterization engines determine the position of the sampling point first. Above that, Cycles implemented global illumination calculation, which resulted in more realistic rendered images \cite{lan2016development}.
Cycles can run on CPUs, while GPU is used to accelerate the rendering process.

\begin{figure}[!ht] \centering
\includegraphics[width=0.23\textwidth]{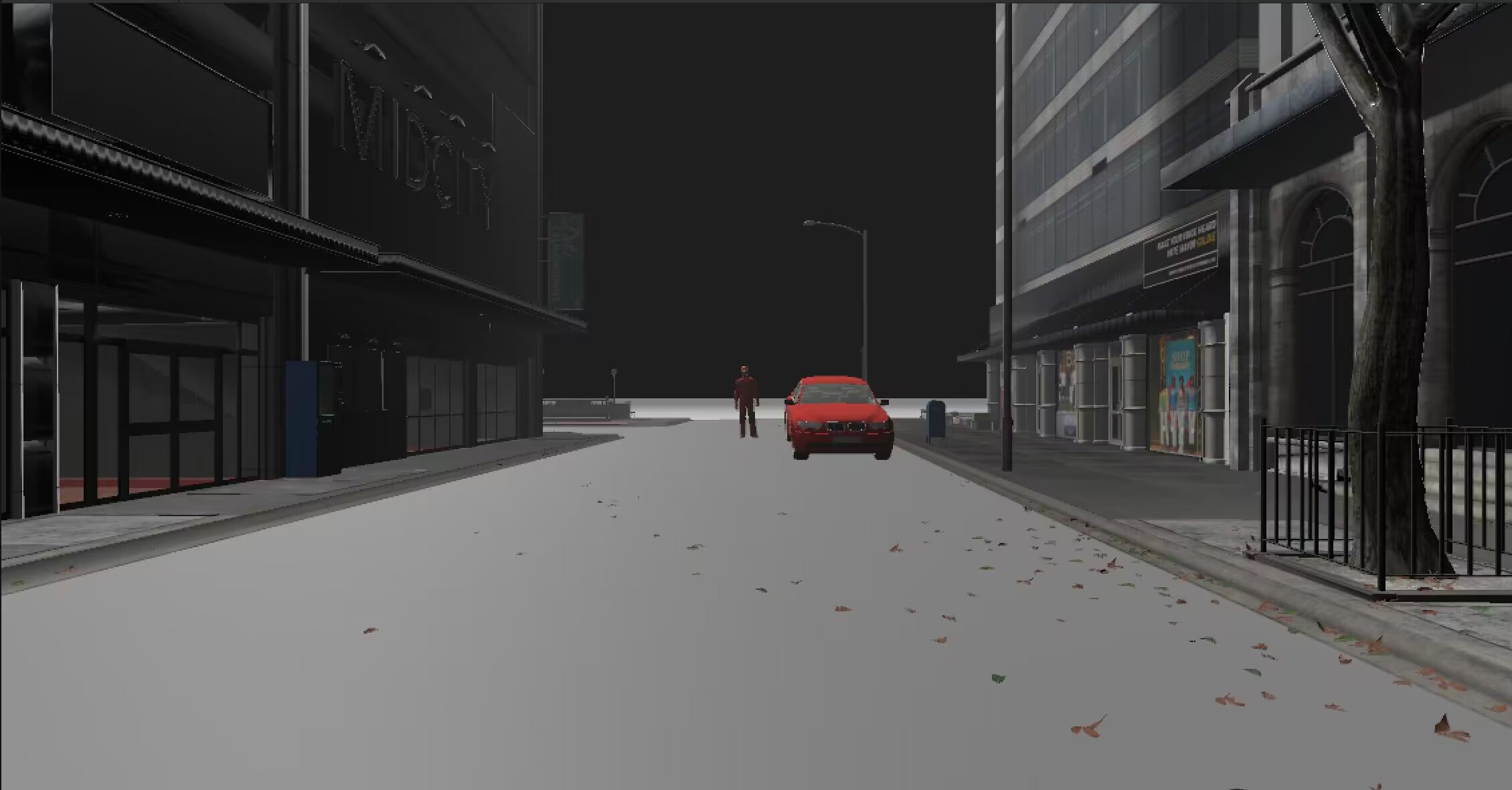}
\includegraphics[width=0.23\textwidth]{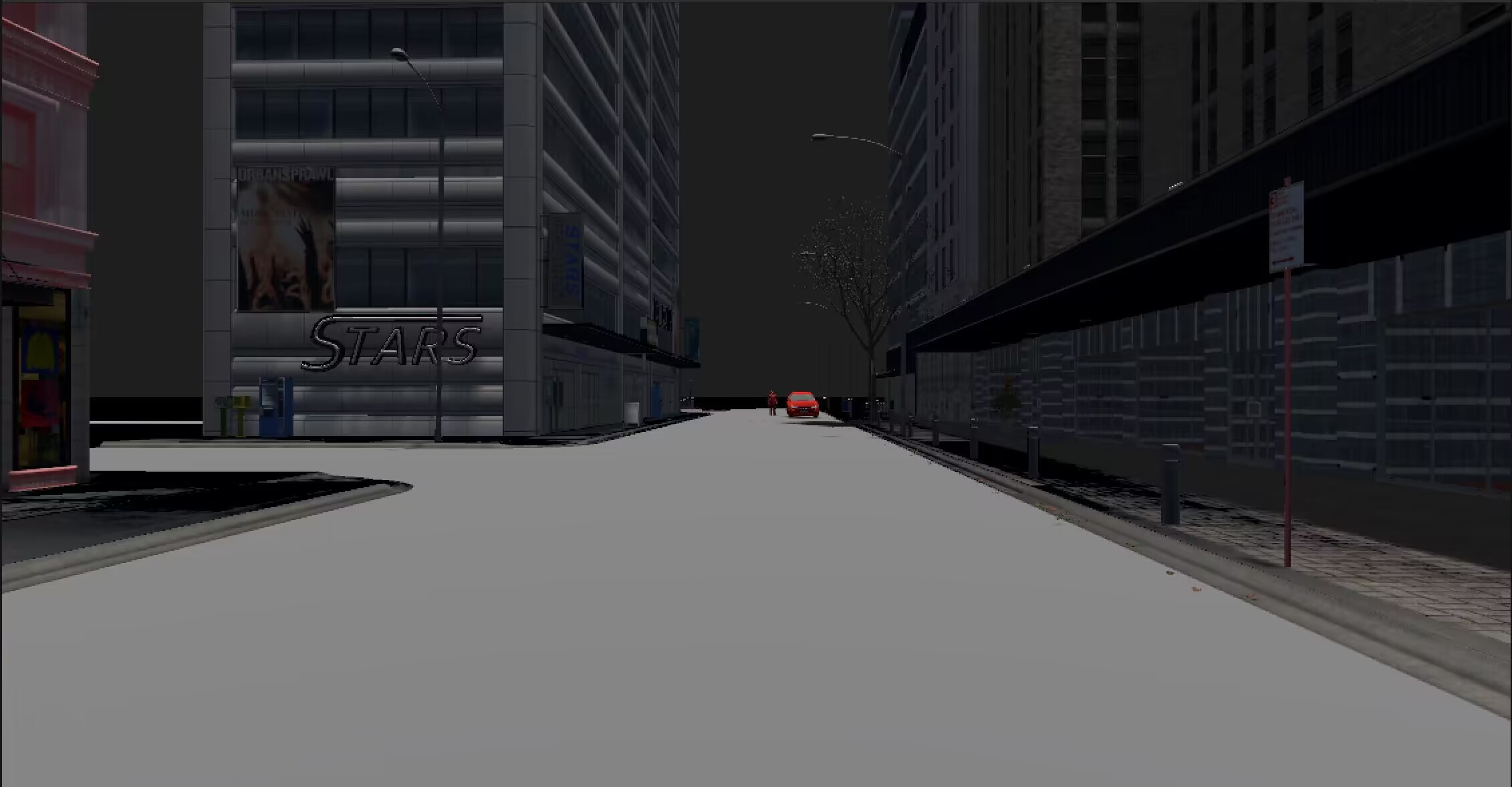}
\includegraphics[width=0.23\textwidth]{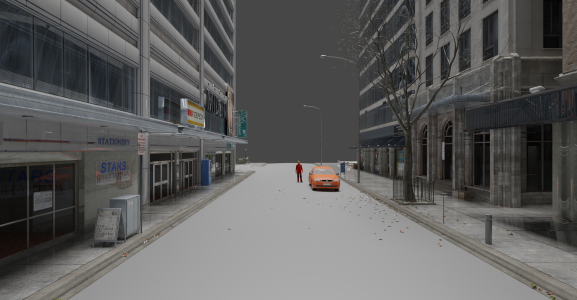}
\includegraphics[width=0.23\textwidth]{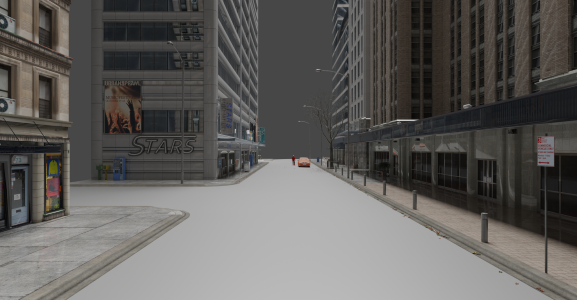}
\caption{Images of similar scenes in both online(above) and offline(below) engines.} 
\end{figure} 

\subsection{Evaluation}
We will use several methods of image quality evaluation mentioned earlier to compare the quality of images generated by offline rendering with those generated by online rendering. We will analyze in which aspects the quality of offline-rendered images is better than that of online-rendered images, and also note in which aspects the quality of images has not significantly improved. We hope to improve our offline rendering through this analysis process.

To quantitatively and visually evaluate the effectiveness of our optimization method \cite{lan2022time}, we have decided to design the following experiment:

We will perform real-time rendering and offline rendering on a set of identical 3D models. And then we capture scene images with the camera positioned at the same location after real-time rendering and offline rendering.
We will apply the object detection algorithm to detect objects (such as pedestrians and vehicles) relevant to driving in both data sets. Finally, we will compare the recognition performance (confidence scores) of these objects in the two sets of images.

Specifically, we set two locations for the main camera to provide the close-up perspective and long-distance perspective of the objects.

\begin{figure}[!ht] \centering
\includegraphics[width=0.48\textwidth]{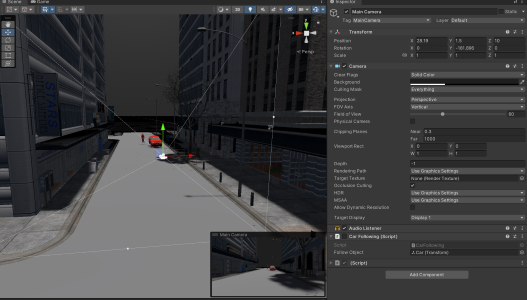}
\caption{close-up perspective}
\end{figure} 

\begin{figure}[!ht] \centering
\includegraphics[width=0.48\textwidth]{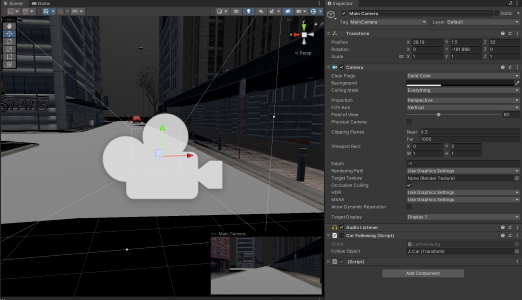}
\caption{long-distance perspective and offline(below) engines.} 
\end{figure} 

The confidence score on the objects using Yolov8n and Detectron2 are shown in \autoref{tab:yolo} and \autoref{tab:detectron} respectively.


\begin{table}[!ht]  \centering 
    \begin{tabular}{l|l|l} \toprule
        \renewcommand{\arraystretch}{1.0} \setlength\tabcolsep{6pt}
         YOLO & Close-up perspective & Long-distance perspective \\ \midrule
         real-time rendering & \makecell[c]{Person:71\% \\ Car:82\%} & \makecell[c]{Person: $<$ 25\% \\ Car: $<$ 25\%} \\  \midrule
         Offline-rendering & \makecell[c]{Person: 88.5\% \\ Car:72\%} & \makecell[c]{Person:42\% \\ Car:34\%} \\ \bottomrule
    \end{tabular}
    \caption{YOLO results.}  \label{tab:yolo}
\end{table}

\begin{table}[!ht]  \centering 
    \begin{tabular}{l|l|l} \toprule
        \renewcommand{\arraystretch}{1.0} \setlength\tabcolsep{3pt}
         Detectron2 & Close-up perspective & Long-distance perspective \\ \midrule
         real-time rendering & \makecell[c]{Person:100\% \\ Car:99\%} & \makecell[c]{Person:68\% \\ Car:90\%} \\ \midrule
         Offline-rendering & \makecell[c]{Person: 100\% \\ Car:98\%} & \makecell[c]{Person:99\% \\ Car:83\%} \\ \bottomrule
    \end{tabular}
    \caption{Detectron results.}  \label{tab:detectron}
\end{table}

While performing better under the object detector does not necessarily guarantee the effectiveness of our optimization method at all times, it remains an important metric indicating that our optimization results exhibit superior computer vision capabilities for autonomous driving compared to the original data.

\section{Results}
\subsection{IQA algorithm}
The calculation results of peak signal-to-noise ratio are shown in \autoref{tab:PSNR}.
\begin{table}[!ht]  \centering 
    \begin{tabular}{c|c|c} \toprule
         & set 1 & set 2 \\ \midrule
         PSNR & 14.015 & 15.425 \\ \bottomrule
    \end{tabular}
    \caption{The results in PSNR.}  \label{tab:PSNR}
\end{table}
In the results of real-time and offline rendering in the near and far groups, the PSNR value of the near group is 14.015, and the PSNR value of the far group is 15.425. This result indicates that the images of the two rendering methods can be identified as similar in terms of visual recognition.

The calculation results of Structural Similarity are shown in \autoref{tab:SSIM}.
\begin{table}[!ht]  \centering 
    \begin{tabular}{c|c|c} \toprule
         & set 1 & set 2 \\ \midrule
         SSIM & 0.5126 & 0.5375 \\ \bottomrule
    \end{tabular}
    \caption{The results in SSIM.}  \label{tab:SSIM}
\end{table}
In the real-time and offline rendering results of the near and far groups, the SSIM values of the near group are 0.5126, and the SSIM values of the far group are 0.5375. This result indicates that the images of the two rendering methods can be considered similar in terms of structural similarity, including brightness, contrast, and structure.

The calculation results of Natural Image Quality Evaluator are shown in \autoref{tab:NIQE}.
\begin{table}[!ht]  \centering 
    \begin{tabular}{l|c|c} \toprule
         NIQE & set 1 & set 2 \\ \midrule
         RealTime & 38.463 & 40.344 \\ 
         Offline & 25.129 & 30.467 \\ \bottomrule
    \end{tabular}
    \caption{The results in NIQE.} \label{tab:NIQE}
\end{table}

In the results of real-time and offline rendering in the near and far groups, the NIQE value for the near group is 38.463, while the NIQE value for offline rendering is 25.129. The NIQE value for the far group is 40.344, while the NIQE value for offline rendering is 30.467. This result indicates that through the evaluation method of NIQE, the quality of offline rendered images is higher than that of real-time rendered images.

\section{Conclusion}
Our research system has been successfully constructed. The research methods should be able to effectively prove the effectiveness of improvement from offline rendering for the optimization of automatic driving model training. However, we still lack a more rigorous theoretical basis, general experiment and a definite conclusion.
As a result, the next task of our research group is to further expand in these three directions.

Judging from the results, our current research cannot guarantee that our expectations must be correct. But if we finally succeed in falsification, it can also play a guiding role in the improvement of automatic driving model training.

In the future, we will extend this work with the AI technologies, such as Knowledge graphs \cite{liu2022towards,lan2022semantic}, and optimization algorithms \cite{lan2021learning,lan2021learning2}. 
\bibliographystyle{IEEEtran}
\bibliography{ref}

\end{document}